%% file: main.tex
\documentclass[lettersize,journal]{IEEEtran}
\usepackage{amsmath,amsfonts}
\usepackage{algorithmic}
\usepackage{array}
\usepackage{textcomp}
\usepackage{cite}
\usepackage{stfloats}
\usepackage{url}
\usepackage{verbatim}
\usepackage{graphicx}
\def\BibTeX{{\rm B\kern-.05em{\sc i\kern-.025em b}\kern-.08em
    T\kern-.1667em\lower.7ex\hbox{E}\kern-.125emX}}
\usepackage{balance}
\usepackage{xcolor}
\definecolor{Green}{RGB}{76, 168, 90}
\newcommand{\benchname}{\texttt{LRR-Bench}\xspace}
\usepackage[hidelinks,colorlinks=true]{hyperref}
\usepackage[nameinlink, capitalize]{cleveref}
\usepackage{xspace}
\usepackage{adjustbox}
\usepackage{pifont}
\usepackage{booktabs}
\usepackage{subcaption}
\usepackage{dsfont} 
\begin{document}
\title{LRR-Bench: Left, Right or Rotate? Vision-Language models Still Struggle With Spatial Understanding Tasks}
\author{Fei Kong, Jinhao Duan, Kaidi Xu, Zhenhua Guo, Xiaofeng Zhu, Xiaoshuang Shi
\thanks{F. Kong, X. Zhu, X. Shi are with  University of Electronic
Science and Technology of China, J. Duan, K. Xu are with Drexel University, Zhenhua Guo is with Tianyijiaotong Technology Ltd., China.}
\thanks{Co-first authors: F. Kong (kong13661@outlook.com), J. Duan}
\thanks{Corresponding author: X. Shi (xsshi2013@gmail.com)}}



\maketitle

\input{sections/abstract}


    \input{sections/introduction.tex}
    \input{sections/related_work.tex}

\input{sections/method.tex}
    \input{sections/experiment.tex}
    \input{sections/conclusion.tex}



\bibliographystyle{IEEEtran}



\end{document}

%% file: sections/abstract.tex
\begin{abstract}
Real-world applications, such as autonomous driving and humanoid robot manipulation, require precise spatial perception. However, it remains underexplored how Vision-Language Models (VLMs) recognize spatial relationships and perceive spatial movement. In this work, we introduce a spatial evaluation pipeline and construct a corresponding benchmark. Specifically, we categorize spatial understanding into two main types: absolute spatial understanding, which involves querying the absolute spatial position (e.g., left, right) of an object within an image, and 3D spatial understanding, which includes movement and rotation. Notably, our dataset is entirely synthetic, enabling the generation of test samples at a low cost while also preventing dataset contamination. We conduct experiments on multiple state-of-the-art VLMs and observe that there is significant room for improvement in their spatial understanding abilities. Explicitly, in our experiments, humans achieve near-perfect performance on all tasks, whereas current VLMs attain human-level performance only on the two simplest tasks. For the remaining tasks, the performance of VLMs is distinctly lower than that of humans. In fact, the best-performing Vision-Language Models even achieve near-zero scores on multiple tasks. The dataset and code are available on \url{https://github.com/kong13661/LRR-Bench}.
\end{abstract}

%% file: sections/introduction.tex
\section{Introduction}
Spatial understanding~\cite{yi2019clevrer} is the capacity of a system to
accurately perceive and interpret the arrangement of objects, including their positions,
orientations, and movements in a given environment. In safety-critical
applications such as robotic control~\cite{brohan2023rt,zitkovich2023rt,kim2024openvla}
and autonomous driving~\cite{zhou2024vision}, Large Vision-Language Models (VLMs)~\cite{zhang2024vision}
depend on this precise spatial awareness to effectively navigate complex environments
and ensure reliable, safe operations.
Existing benchmarks~\cite{liu2023visual,shiri2024empirical,xie2024expand,du2024embspatial}
for spatial understanding primarily focus on inferring spatial relationships in natural
images by formulating queries about the relative positions of two objects (e.g.,
left, in front of). These benchmarks typically leverage auxiliary techniques, such as depth estimation and segmentation~\cite{kirillov2023segment} to annotate spatial
relationships automatically. Additionally, synthetic datasets, including grid-based
images~\cite{wu2024vsp, wang2025picture, tang2024sparkle,tang2024grasp} or
psychrometric charts~\cite{ xu2025defining}, have been utilized to examine model
behaviors and compare them against human perception.

However, current benchmark queries primarily focus on basic spatial
relationships, e.g., position. More realistic and complex spatial relationships
and capabilities, such as motion perception, relative movement, and sequential movement,
remain largely unexplored. In this paper, we provide a spatial understanding
benchmark, \benchname, to comprehensively evaluate the spatial understanding
capabilities of VLMs from two perspectives \textit{absolute position} and \textit{relative
movement}, in both 2D and 3D scenarios. For absolute positioning scenarios, we perform
spatial relationship inference in both static and dynamic contexts. Specifically,
we query positional relationships between objects in static scenarios and analyze
directions of motion or movement in dynamic image sequences. Regarding relative
movement, we consider scenarios involving motion caused by both object and camera
movements, which presents challenges to the fundamental spatial reasoning
abilities of VLMs.

Building upon \benchname, we perform extensive experiments on 20+ state-of-the-art
VLMs, including both commercial~\cite{achiam2023gpt} and open-source models with
diverse parameter sizes (up to 72B for open-source VLMs) and varied training
strategies, such as those with or without preference optimization~\cite{ouyang2022training} and finetuning with 3D dataset.
We find that existing VLMs perform well on simple absolute positional
relationship inference but struggle significantly with image sequences and
understanding relative movements. By contrast, our human evaluation
demonstrates that humans effortlessly achieve around 90\% accuracy on \benchname,
whereas VLMs perform close to random guessing in most cases.

\begin{figure*}[t!]
    \centering
    \begin{subfigure}
        [t]{0.95\textwidth}
        \centering
        \includegraphics[width=1\textwidth]{
            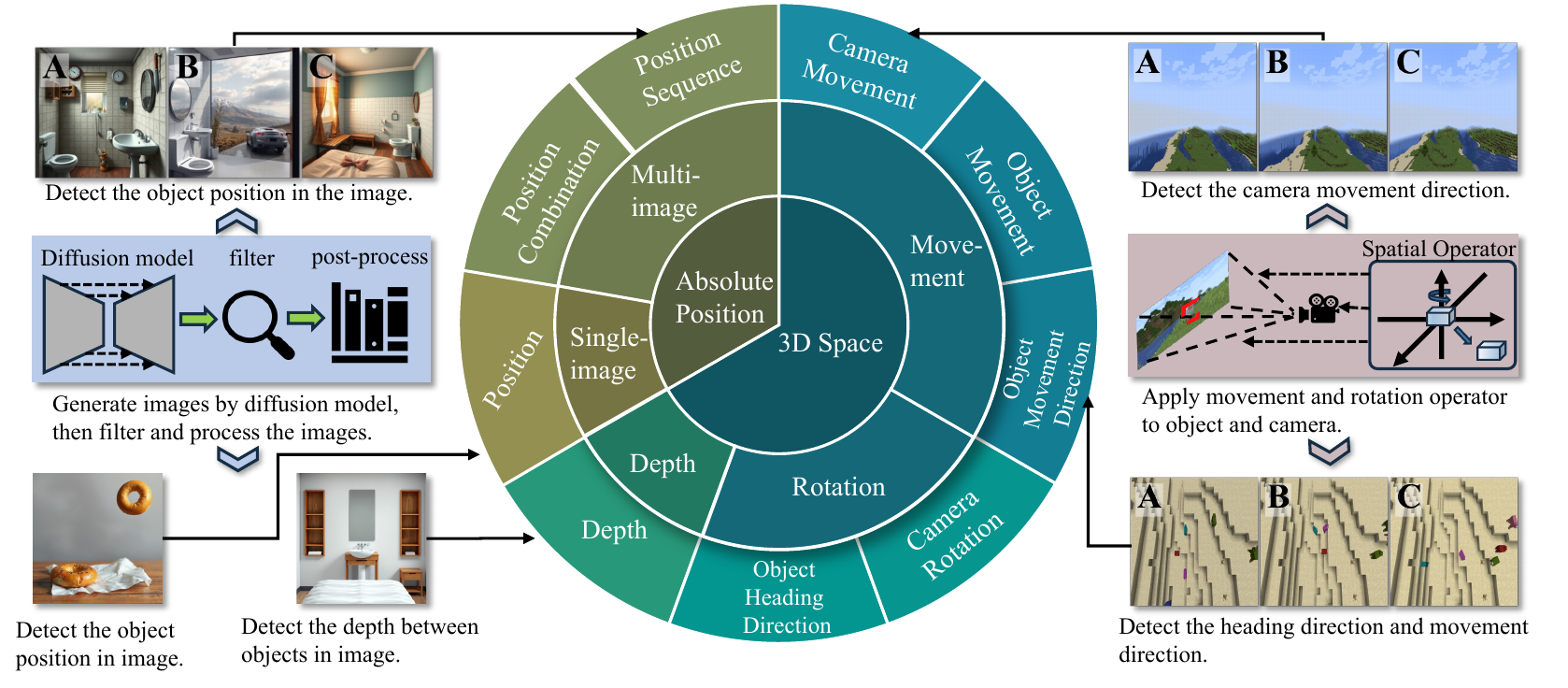
        } 
    \end{subfigure}
    \caption{This diagram illustrates our categorization of the spatial
    understanding problem and the overall pipeline. The blue section represents
    3D spatial understanding, while the yellow section represents absolute
    position understanding. In 3D spatial understanding, we decompose spatial
    translation into rotation and movement, applying them separately to the camera
    and the object. In absolute position understanding, we detect the object's absolute
    position within the image, such as the center, top-left, and so on. For the tasks
    related to absolute position and depth, the samples are generated using a
    diffusion model. These samples are then filtered by GroundingDINO and processed
    by various models. Samples for the other tasks are generated by applying
    movement and rotation to the camera and object within Minecraft.}
    \label{fig:pipeline}
\end{figure*}

Apart from qualitative results, we also observe that \ding{202} advanced
reasoning methods, e.g., CoT~\cite{wei2022chain}, do not consistently improve
spatial understanding; \ding{203} preference optimization, e.g., Mixed Preference
Optimization (MPO)~\cite{wang2024enhancing}, can negatively impact spatial
understanding; \ding{204} parameter scaling laws are ineffective for enhancing
spatial understanding. These insights highlight that spatial understanding is distinct
from common reasoning benchmarks. We hope our \benchname will advance research
and development in this area. Our contributions can be summarized as follows:
\begin{itemize}
    \item We propose a fully synthetic spatial reasoning dataset and a corresponding
        pipeline, which enables the low-cost generation of test tasks. By utilizing
        fully synthetic test samples, our approach effectively prevents dataset
        contamination.

    \item We evaluate the absolute spatial understanding and the 3D spatial understanding.
        For 3D spatial understanding, we decompose it into rotation and movement
        and test on camera, object perspective.

    \item We evaluate state-of-the-art models and find that there is still significant
        room for improvement in their spatial understanding and reasoning will
        induce hallucination for some tasks.
\end{itemize}

%% file: sections/related_work.tex
\section{Related Work}

\subsection{Large Vision-Language Models}
Large Vision-Language Models (VLMs)~\cite{zhang2024vision} are a type of language
models enhanced with visual adapters, enabling them to process and interpret
visual inputs, such as images, and generate responses based on given
instructions~\cite{ouyang2022training}. There have been a variety of LVLM families
proposed out of various purposes and designs, such as BLIP~\cite{li2022blip,li2023blip},
LLaVA~\cite{liu2023visual,liu2024llava}, mPlug-Owl~\cite{ye2023mplug,ye2024mplug},
MiniGPT-4~\cite{zhu2023minigpt}, QWenVL~\cite{wang2024qwen2,bai2025qwen2}, InternVL~\cite{chen2024internvl}.
For instance, BLIP~\cite{li2022blip} utilizes a unified approach to vision-language
modeling by jointly training visual understanding and language generation tasks to
enhance image descriptions. Building on this, BLIP2~\cite{li2023blip} is introduced
to further harness the capabilities of LLMs by incorporating a frozen image encoder.
Other VLMs follow a similar paradigm to BLIP-2, where a visual query adapter is designed
to enhance the language model’s visual understanding. However, they differ in training
methods, visual adapter design, and training data. For example, the LLaVA familyLlava~\cite{liu2023visual,liu2024llava}
leverages CLIP~\cite{radford2021learning} as its vision encoder and employs a Q-former~\cite{li2023blip}
training strategy to align vision embeddings with LLaMA, making it particularly
strong in instruction-following and open-ended tasks. By contrast, MiniGPT-4~\cite{zhu2023minigpt}
integrates a pre-trained BLIP-2 vision encoder and applies low-rank adaptation (LoRA)~\cite{hu2022lora}
to efficiently fine-tune the language model. Recent advanced VLMs, such as QWen~\cite{wang2024qwen2,bai2025qwen2}
and InternVL~\cite{chen2024internvl}, have been further improved through
the use of higher-quality training data, including multi-turn dialogues and web-scale
image-text pairs, enhancing their alignment with generic visual-linguistic tasks.

\begin{figure*}[t!]
    \centering
    \begin{subfigure}
        [t]{0.95\textwidth}
        \centering
        \includegraphics[width=1\textwidth]{
            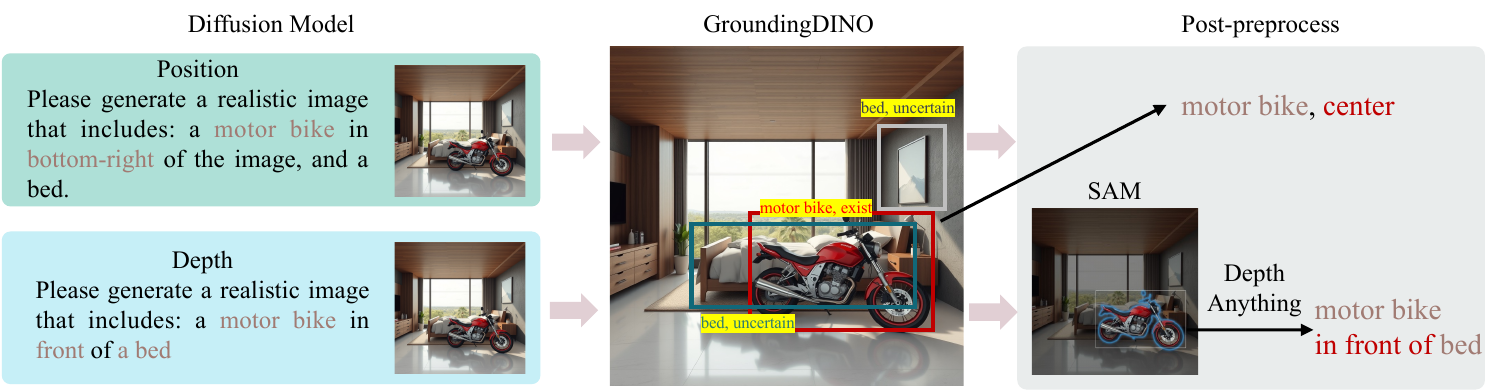
        } 
    \end{subfigure}
    \caption{This diagram illustrates the process of generating samples using
    the diffusion model. The first column displays the prompt. The generated
    samples are then fed to GroundingDINO, which outputs the bounding box and a confidence
    score. The confidence score is categorized into three classes: existing classes,
    non-existing classes, and uncertain classes. The bounding boxes and
    confidence scores are used to filter the samples. The filtered samples are subsequently
    fed to the next stage, which varies depending on the specific task.}
    \label{fig:pipeline2}
\end{figure*}

There are also several studies have explored ways to enhance the spatial understanding
capabilities of foundation models. SpatialCoT~\cite{liu2025spatialcot} improves spatial
reasoning through coordinate alignment and Chain-of-Thought (CoT) for planning.
SALE~\cite{tao2024enhancing} enhances spatial reasoning by incorporating
structured textual descriptions. SpatialBot~\cite{cai2024spatialbot} extracts depth
information of target objects from depth maps and then reasons about their spatial
relationships. SpaceQwen25~\cite{spaceqwen}, SpaceOM~\cite{spaceom}, Llava-3D~\cite{zhu2024llava} utilize the synthetic 3D dataset to enhance the 3D spatial understanding.
\subsection{Spatial Understanding in VLMs}
Spatial understanding is crucial in safety-critical applications such as robotic
control~\cite{zitkovich2023rt,kim2024openvla} and autonomous driving~\cite{zhou2024vision}.
To assess the spatial understanding of VLMs, VSR~\cite{liu2023visual}, What'sUP~\cite{kamath2023s} and Spatial-MM~\cite{shiri2024empirical}
generate spatial relation queries (e.g., under, in front of, facing) from natural
images. VSRE~\cite{xie2024expand} extends VSR by refining both data and model
tuning to address its imbalanced sensitivity to instructions, where VSR is overly
sensitive to language cues but under-responsive to visual positional information.
By contrast, EmbSpatial~\cite{du2024embspatial}, 3DSRBench~\cite{ma20243dsrbench}, COMFORT~\cite{zhang2024vision_bench}, and LEFT~\cite{hsu2023s} formulates spatial relation queries
from 3D scenes, emphasizing embodied intelligence. However, these benchmarks
primarily focus on positional relationships and offer a limited scope of spatial
understanding, overlooking aspects such as camera movement and \textbf{continuous visual
signal analysis}, which are crucial for real-world applications. In~\cref{tab:taxonomy}, we compare \benchname with previous spatial
understanding benchmarks.


\begin{table}[]
    \caption{Taxonomy of existing spatial understanding benchmarks. ``Nat. Img.''
    refers to natural images and ``Absolute Pos.'' refers to absolute position.}
    \centering
    \adjustbox{width=\linewidth}{
    \begin{tabular}{lcccccc}
        \toprule \textbf{Benchmark} & \textbf{Nat. Img.}                    & \textbf{Absolute Pos.}            & \textbf{Relative Move}                & \textbf{Camera Move}                  & \textbf{3D}                           & \textbf{Sequence}                     \\
        \toprule VSR                & $\Large\color{Green}{\checkmark}$     & $\Large\color{Green}{\checkmark}$ & $\Large\color{red}{\pmb{\mathsf{X}}}$ & $\Large\color{red}{\pmb{\mathsf{X}}}$ & $\Large\color{red}{\pmb{\mathsf{X}}}$ & $\Large\color{red}{\pmb{\mathsf{X}}}$ \\
        Spatial-MM                  & $\Large\color{Green}{\checkmark}$     & $\Large\color{Green}{\checkmark}$ & $\Large\color{red}{\pmb{\mathsf{X}}}$ & $\Large\color{red}{\pmb{\mathsf{X}}}$ & $\Large\color{red}{\pmb{\mathsf{X}}}$ & $\Large\color{red}{\pmb{\mathsf{X}}}$ \\
        What'sUP & $\Large\color{Green}{\checkmark}$ & $\Large\color{Green}{\checkmark}$ & $\Large\color{red}{\pmb{\mathsf{X}}}$ & $\Large\color{red}{\pmb{\mathsf{X}}}$ & $\Large\color{red}{\pmb{\mathsf{X}}}$ & $\Large\color{red}{\pmb{\mathsf{X}}}$ \\
        VSRE                        & $\Large\color{Green}{\checkmark}$     & $\Large\color{Green}{\checkmark}$ & $\Large\color{red}{\pmb{\mathsf{X}}}$ & $\Large\color{red}{\pmb{\mathsf{X}}}$ & $\Large\color{red}{\pmb{\mathsf{X}}}$ & $\Large\color{red}{\pmb{\mathsf{X}}}$ \\
        EmbSpatial                  & $\Large\color{red}{\pmb{\mathsf{X}}}$ & $\Large\color{Green}{\checkmark}$ & $\Large\color{red}{\pmb{\mathsf{X}}}$ & $\Large\color{red}{\pmb{\mathsf{X}}}$ & $\Large\color{red}{\pmb{\mathsf{X}}}$ & $\Large\color{red}{\pmb{\mathsf{X}}}$ \\
        Sparkle                     & $\Large\color{Green}{\checkmark}$     & $\Large\color{Green}{\checkmark}$ & $\Large\color{Green}{\checkmark}$     & $\Large\color{red}{\pmb{\mathsf{X}}}$ & $\Large\color{red}{\pmb{\mathsf{X}}}$ & $\Large\color{red}{\pmb{\mathsf{X}}}$ \\
        Grasp                       & $\Large\color{red}{\pmb{\mathsf{X}}}$ & $\Large\color{Green}{\checkmark}$ & $\Large\color{Green}{\checkmark}$     & $\Large\color{red}{\pmb{\mathsf{X}}}$ & $\Large\color{red}{\pmb{\mathsf{X}}}$ & $\Large\color{red}{\pmb{\mathsf{X}}}$ \\
        VSP                         & $\Large\color{red}{\pmb{\mathsf{X}}}$ & $\Large\color{Green}{\checkmark}$ & $\Large\color{Green}{\checkmark}$     & $\Large\color{red}{\pmb{\mathsf{X}}}$ & $\Large\color{red}{\pmb{\mathsf{X}}}$ & $\Large\color{red}{\pmb{\mathsf{X}}}$ \\
        BSAs                        & $\Large\color{red}{\pmb{\mathsf{X}}}$ & $\Large\color{Green}{\checkmark}$ & $\Large\color{Green}{\checkmark}$     & $\Large\color{red}{\pmb{\mathsf{X}}}$ & 
        $\Large\color{Green}{\checkmark}$     & $\Large\color{red}{\pmb{\mathsf{X}}}$ \\
        3DSRBench & $\Large\color{Green}{\checkmark}$ & $\Large\color{Green}{\checkmark}$ & $\Large\color{Green}{\checkmark}$ & $\Large\color{red}{\pmb{\mathsf{X}}}$ & $\Large\color{Green}{\checkmark}$ & $\Large\color{red}{\pmb{\mathsf{X}}}$ \\
        LEFT & $\Large\color{Green}{\checkmark}$ & $\Large\color{Green}{\checkmark}$ & $\Large\color{red}{\pmb{\mathsf{X}}}$ & $\Large\color{red}{\pmb{\mathsf{X}}}$ & $\Large\color{Green}{\checkmark}$ & $\Large\color{red}{\pmb{\mathsf{X}}}$ \\
        COMFORT & $\Large\color{red}{\pmb{\mathsf{X}}}$ & $\Large\color{Green}{\checkmark}$ & $\Large\color{red}{\pmb{\mathsf{X}}}$ & $\Large\color{red}{\pmb{\mathsf{X}}}$ & $\Large\color{Green}{\checkmark}$ & $\Large\color{red}{\pmb{\mathsf{X}}}$ \\
        \midrule \benchname         & $\Large\color{Green}{\checkmark}$     & $\Large\color{Green}{\checkmark}$ & $\Large\color{Green}{\checkmark}$     & $\Large\color{Green}{\checkmark}$     & $\Large\color{Green}{\checkmark}$     & $\Large\color{Green}{\checkmark}$     \\ 
        \bottomrule
    \end{tabular}
    }
    \label{tab:taxonomy}
\end{table}

%% file: sections/method.tex
\section{Data collection}

\cref{fig:pipeline} illustrates the pipeline and categorization of our dataset.
Specifically, we divide spatial understanding into 3D spatial understanding and absolute
position understanding. Our data is generated using generative models, e.g., diffusion models, or
Minecraft, so as to reduce the cost of the dataset construction and facilitates easy expansion of
our pipeline. A detailed description
of the pipeline will be provided in the following two sections.

\begin{figure}[t]
    \centering
    \begin{subfigure}
        [t]{0.2\textwidth}
        \centering
        \includegraphics[width=0.7\textwidth]{
            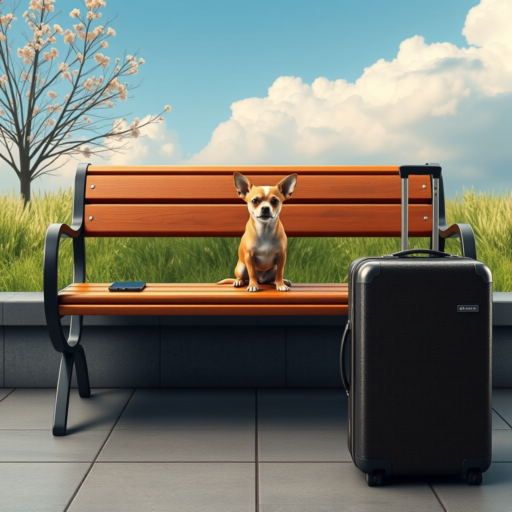
        } 
        \captionsetup{skip=3pt, width=1\linewidth, justification=centering}
        \subcaption[]{\textcolor{olive}{The negative sample.}}
    \end{subfigure}
    \begin{subfigure}
        [t] {0.2\textwidth}
        \centering
        \includegraphics[width=0.7\textwidth]{
            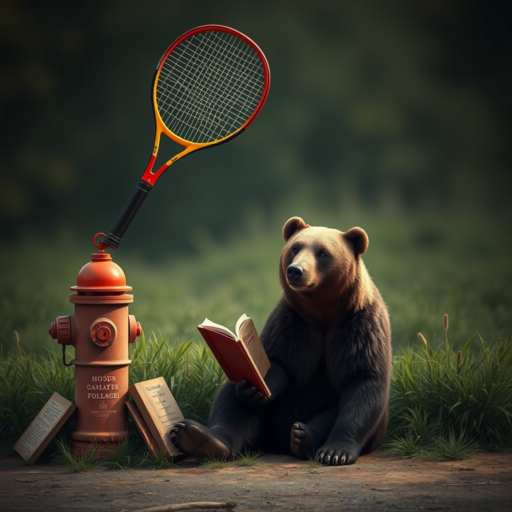
        }
        \captionsetup{skip=3pt, width=1\linewidth} \subcaption[]{\textcolor{orange}{The positive sample.}}
    \end{subfigure}
    \captionsetup{skip=3pt}
    \caption{Please answer if the image has \textcolor{orange}{book} (\textcolor{olive}{suitcase})
    at \textcolor{orange}{bottom-left} (\textcolor{olive}{bottom-left}) of the
    image. Please answer Yes or No.}
    \label{img:position}
\end{figure}

\subsection{Absolute Position Reasoning}
\label{diffusion} We define absolute position reasoning as problems like determining
whether an object is located in the bottom-left corner of the image, where the
answer is independent of the object's position within the scene background. To tackle
this, we first use a diffusion model to initiate image generation and then leverage auxiliary models to filter low-quality images to construct the basic Position (Pos.) task. Building upon this task, we
then create two more challenging tasks: Position Combination (Pos. C.) and Position Sequence (Pos. S.).

\begin{figure}[t]
    \centering
    \begin{subfigure}
        [t]{0.2\textwidth}
        \centering
        \includegraphics[width=1\textwidth]{
            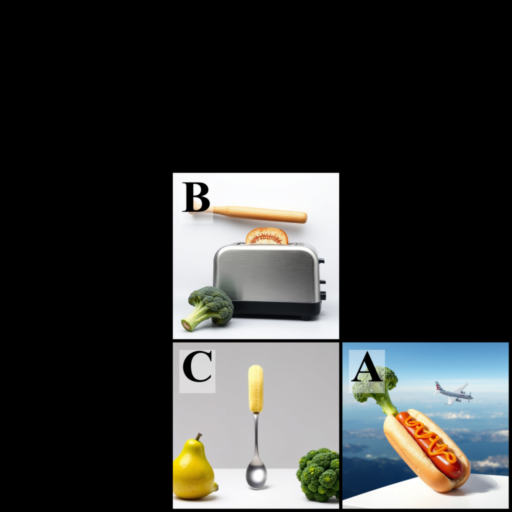
        } 
        \captionsetup{skip=3pt, width=1\linewidth, justification=centering}
        \subcaption[]{\textcolor{olive}{The negative sample.}}
    \end{subfigure}
    \hspace{3mm}
    \begin{subfigure}
        [t] {0.2\textwidth} \captionsetup{skip=3pt, width=1\linewidth}
        \centering
        \includegraphics[width=1\textwidth]{
            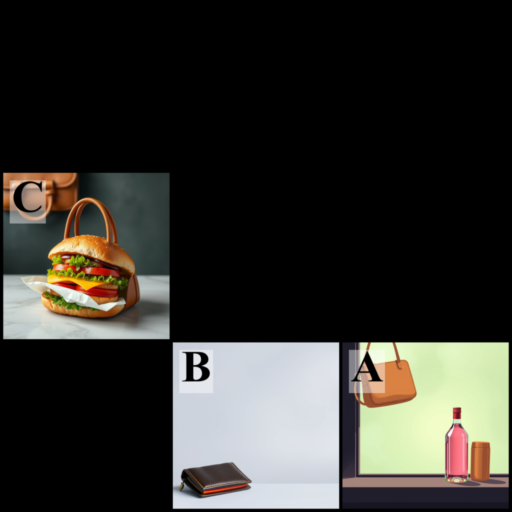
        }
        \subcaption[]{\textcolor{orange}{The positive sample.}}
    \end{subfigure}
    \captionsetup{skip=3pt}
    \caption{Follow the subplot order A, B, C to check whether there is a
    \textcolor{orange}{handbag} (\textcolor{olive}{broccoli}) located in each
    subplot matches \textcolor{orange}{top-left} (\textcolor{olive}{top-left}), \textcolor{orange}{bottom-left}
    (\textcolor{olive}{bottom-right}), and \textcolor{orange}{top-left} (\textcolor{olive}{bottom-left}),
    respectively. Please answer Yes if all position is right, otherwise No.}
    \label{img:order}
\end{figure}

\textbf{Position (Pos.)} In this task, we assess the ability of VLMs to
determine whether a specific object is present at a given location within an
image. To generate the necessary samples, we utilize the Flux.1-S model,
prompting it with specific object and location information. However, we observe
that the images generated by Flux.1-S~\cite{labs2025flux1kontextflowmatching,flux2024} do not fully adhere to the provided prompt.
To filter the images effectively and ensure the generalizability of the
filtering model, we apply a zero-shot bounding box predictor, GroundingDINO~\cite{liu2024grounding}, to predict
the bounding boxes for the objects. In the case of the ``existing'' problem, we retain
only those images where the bounding boxes are predicted with high confidence.
For the ``non-existing'' problem, we keep only those images in which the bounding
boxes for the objects have a confidence score above a low threshold.
\cref{fig:pipeline2} illustrates the pipeline. We focus on five specific locations
within the image: \underline{top-left}, \underline{top-right}, \underline{bottom-left}, \underline{bottom-right}, and \underline{center}. These
strategies help ensure high-quality filtered images, which are then fed into the
VLMs for evaluation. This approach allows us to assess the model’s fundamental understanding
of absolute positioning. The samples for this task are shown in
\cref{img:position}.

\textbf{Position Combination (Pos. C.)} In the Position task, the model is
queried about five specific positions, which may not be sufficient to fully evaluate
its understanding of absolute positioning. To better assess this, we introduce a
more challenging task. We combine multiple sub-images into a single composite
image and then inquire about the specific object position within each sub-image
in a given order. To solve this, the model must determine the location of each
sub-image within the larger image, estimate the proportion of space each sub-image
occupies, and identify the position of the object within each sub-image. Furthermore,
the model needs to locate the label associated with each sub-image. This task is
significantly more complex than the basic Position task. In our experiment, we randomly
select three sub-images and combine them into a single image, which is then
divided into nine equal parts. The sub-images are placed in three of these nine parts
in a random order, while the remaining parts are filled with black pixels. Samples
for this task are shown in \cref{img:order}.

\textbf{Position Sequence (Pos. S.)} We also examine VLMs using \underline{image sequences}.
Similar to the Position Combination task, we randomly select three images to
create the sequence. In this sequence, there is no inherent relationship between
the individual images, meaning that changes in one image (e.g., background
variations) do not influence the results of the other images. To correctly
answer the queries posed by this task, the model must isolate the images from one
another, ensuring that no mutual influence exists between them.

\begin{figure}[t]
    \centering
    \begin{subfigure}
        [t]{0.45\textwidth}
        \centering
        \includegraphics[width=1\textwidth]{
            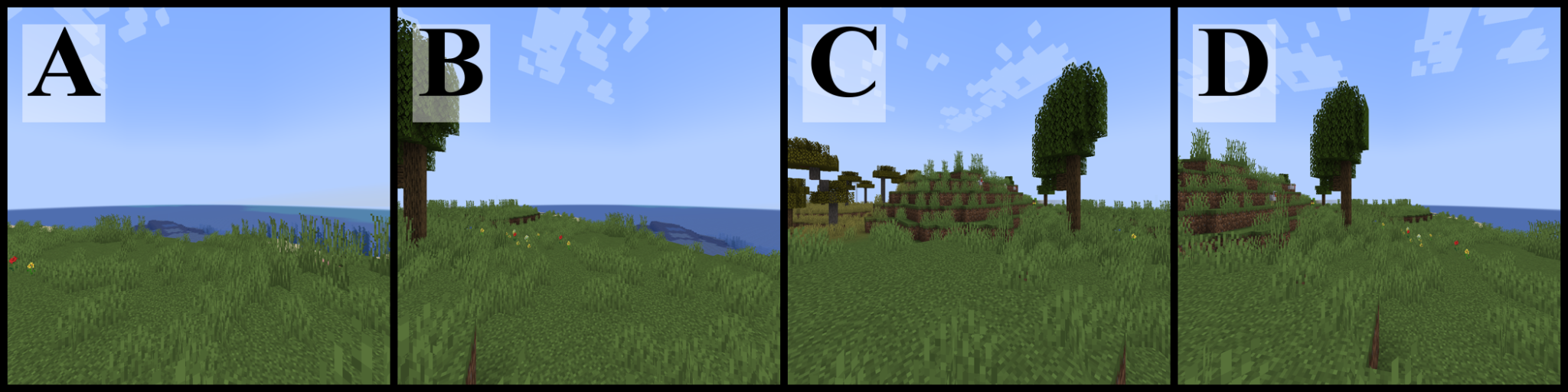
        } 
        \captionsetup{skip=2pt, width=1\linewidth, justification=centering}
        \subcaption[]{\textcolor{olive}{The negative sample.}}
    \end{subfigure}
    \vspace{2mm}

    \begin{subfigure}
        [t] {0.45\textwidth}
        \centering
        \includegraphics[width=1\textwidth]{
            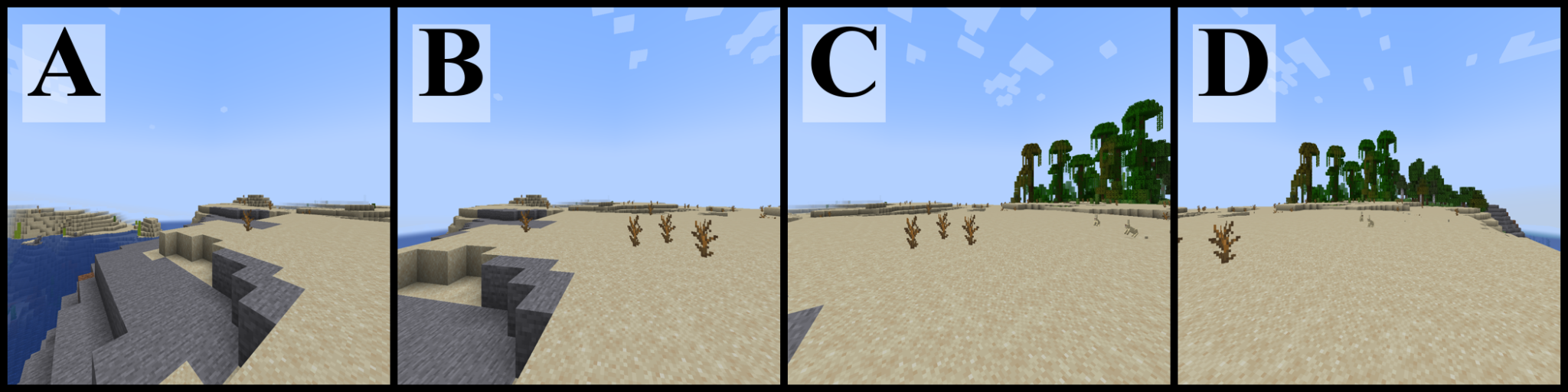
        }
        \captionsetup{skip=3pt, width=1\linewidth} \subcaption[]{\textcolor{orange}{The positive sample.}}
    \end{subfigure}
    \captionsetup{skip=3pt}
    \caption{The background of the sequence is same with different camera. Please answer if the camera's rotatation direction of the image
    sequence is same following A, B, C, D. The answer is either Yes or No. }
    \label{img:rotate}
    \vspace{-4mm}
\end{figure}

\begin{figure}[t]
    \centering
    \begin{subfigure}
        [t]{0.2\textwidth}
        \centering
        \includegraphics[width=0.7\textwidth]{
            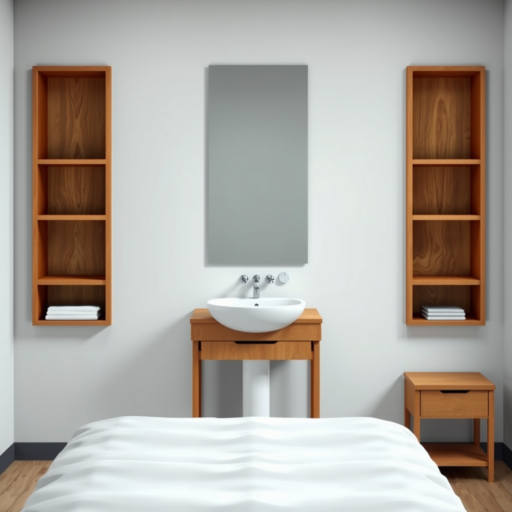
        } 
        \captionsetup{skip=3pt, width=1\linewidth, justification=centering}
        \subcaption[]{\textcolor{olive}{The negative sample.}}
    \end{subfigure}
    \begin{subfigure}
        [t] {0.2\textwidth}
        \centering
        \includegraphics[width=0.7\textwidth]{
            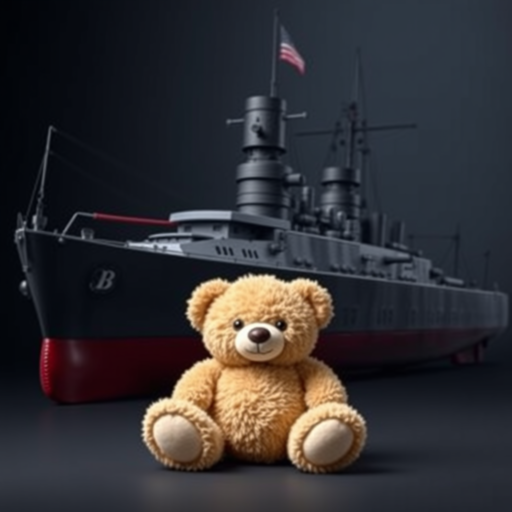
        }
        \captionsetup{skip=3pt, width=1\linewidth} \subcaption[]{\textcolor{orange}{The positive sample.}}
    \end{subfigure}
    \captionsetup{skip=3pt}
    \caption{Please answer if the \textcolor{orange}{teddy bear} (\textcolor{olive}{sink})
    is in front of the \textcolor{orange}{boat} (\textcolor{olive}{bed}). Please
    answer Yes or No.}
    \label{img:depth}
\end{figure}

\subsection{3D Spatial Reasoning}

We define problems like determining whether an object is moving in the background
or whether the camera is rotating as 3D Spatial Reasoning. To answer these types
of questions, VLMs must understand the entire 3D space and the relationships between
objects and that space. Our test samples are constructed from several
perspectives: the relative depth of objects in space, camera rotation, camera movement, object rotation, and object movement.
With the exception of the Depth task, all other tasks require an image sequence. To generate
reliable image sequences at a lower cost, we opt to capture screenshots within
Minecraft to create the image sequences. The following outlines the detailed pipeline.

\textbf{Depth (Dep.)} We first establish a task focused on object depth. Similar
to the Position task, we use the diffusion model Flux.1-S to generate images,
and apply the GroundingDINO model to filter them. After filtering, the bounding
boxes are used as prompts for the SAM~\cite{kirillov2023segment} to obtain object segmentations. The Depth-Anything-V2~\cite{yang2024depth}
model is then utilized to predict the depth of the segmented object areas. Once
the dataset is constructed, we query the VLMs to identify which object is in front.
This task evaluates the model's basic spatial understanding capabilities. Samples
of this task are shown in \cref{img:depth}.

\begin{figure}[t]
    \centering
    \begin{subfigure}
        [t]{0.4\textwidth}
        \centering
        \includegraphics[width=1\textwidth]{
            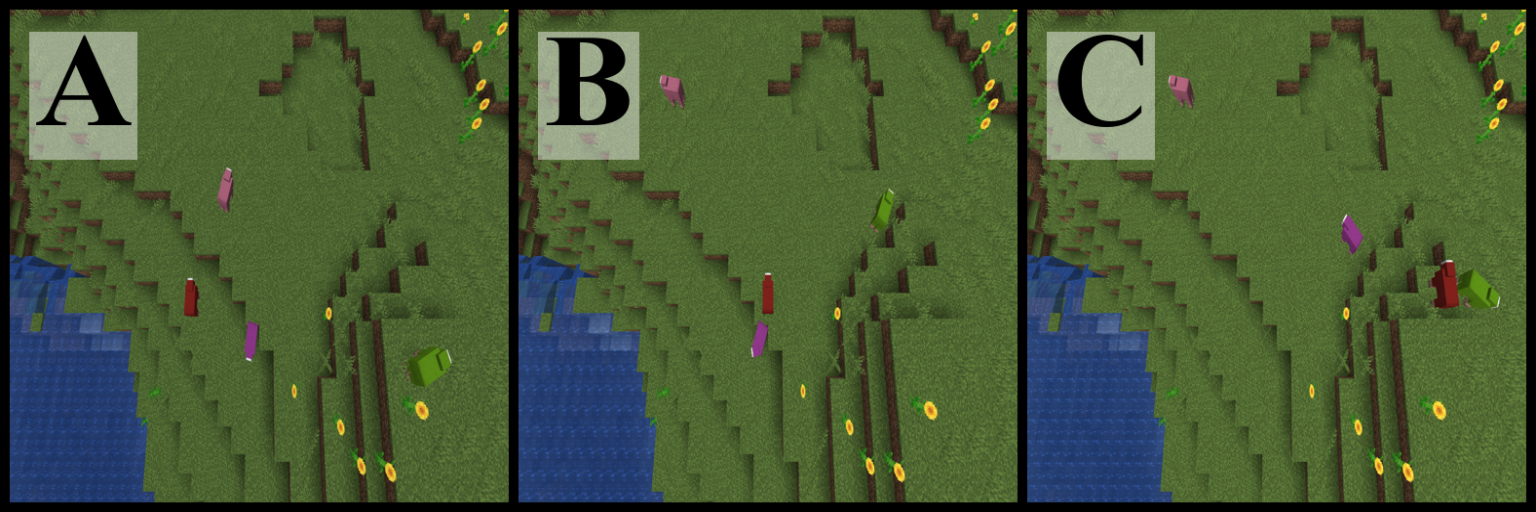
        } 
        \captionsetup{skip=2pt, width=1\linewidth, justification=centering}
        \subcaption[]{\textcolor{olive}{The negative sample.}}
    \end{subfigure}
    \vspace{2mm}

    \begin{subfigure}
        [t] {0.4\textwidth}
        \centering
        \includegraphics[width=1\textwidth]{
            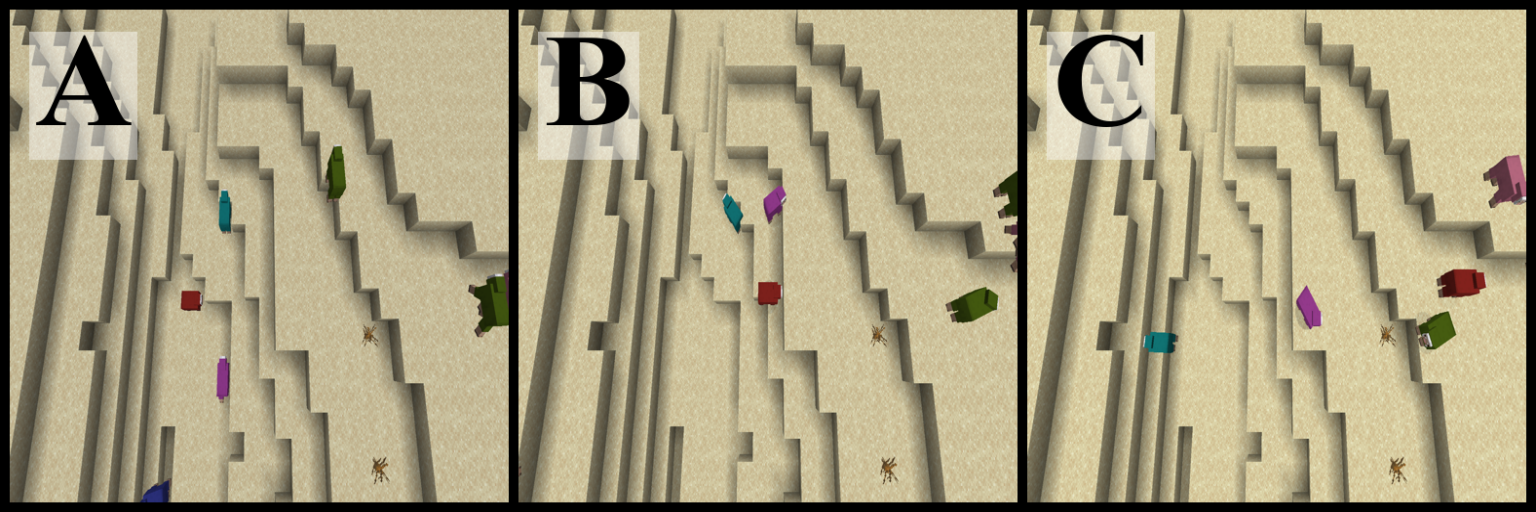
        }
        \captionsetup{skip=3pt, width=1\linewidth} \subcaption[]{\textcolor{orange}{The positive sample.}}
    \end{subfigure}
    \captionsetup{skip=3pt}
    \caption{Each image has a orange sheep (orange block) in different position. The background and camera are fixed. Please answer if the move
    direction of the orange sheep (orange block) is same with the direction of its
    head (white head) following the sequence of A, B, C. The answer is either
    Yes or No.}
    \label{img:sheep_move_direction}
\end{figure}
\textbf{Camera Rotation (Ca. R.)} As shown in \cref{fig:pipeline}, we decompose the
spatial transformation into rotation and movement. This task focuses on the
rotation of the camera. We keep the position of the game camera fixed and rotate
the camera by a fixed angle multiple times. To ensure the view remains
unobstructed, we elevate the camera a certain distance above the ground while capturing
the screenshots. After each rotation, we take a screenshot. Once the image
sequence is obtained, we query the VLMs to determine whether the viewpoint rotation
remains consistent in its direction. For positive samples, we preserve the original
sequence of screenshots. For negative samples, we select two images from the sequence
and swap their positions. Samples for this task are shown in \cref{img:rotate}.

\begin{figure}[t]
    \centering
    \begin{subfigure}
        [t]{0.45\textwidth}
        \centering
        \includegraphics[width=1\textwidth]{
            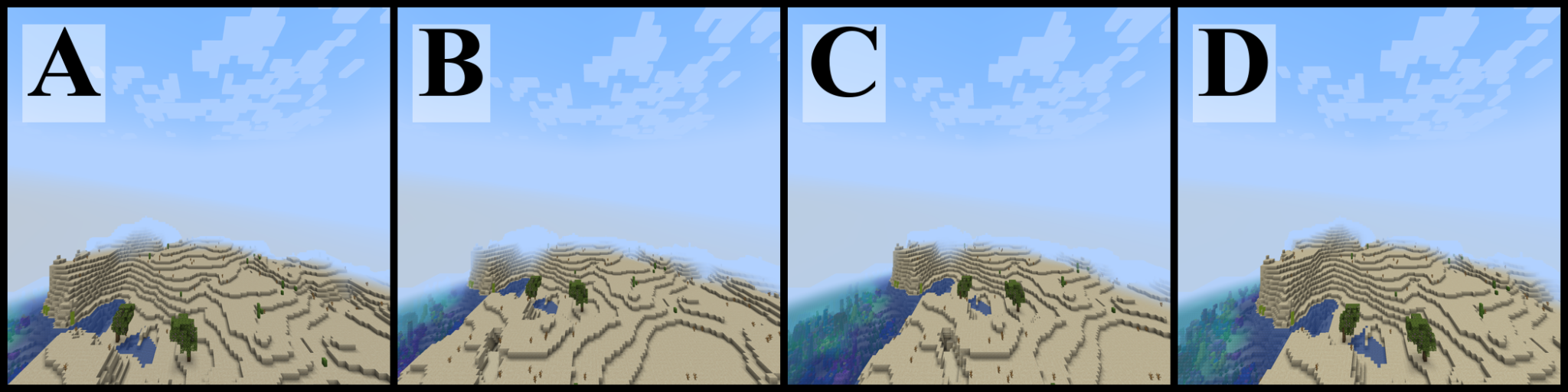
        } 
        \captionsetup{skip=2pt, width=1\linewidth, justification=centering}
        \subcaption[]{\textcolor{olive}{The negative sample.}}
    \end{subfigure}
    \vspace{2.5mm}

    \begin{subfigure}
        [t] {0.45\textwidth}
        \centering
        \includegraphics[width=1\textwidth]{
            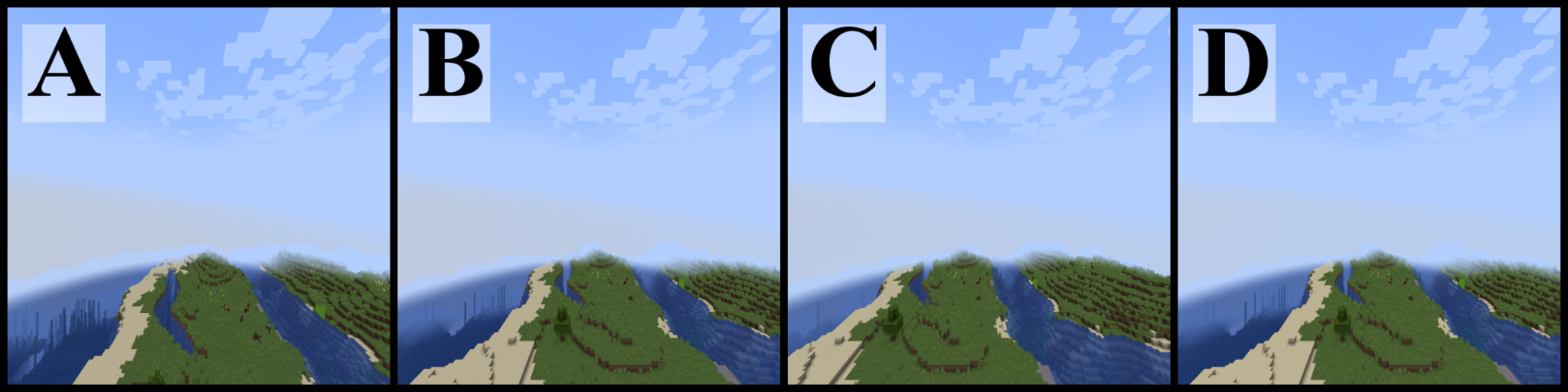
        }
        \captionsetup{skip=2pt, width=1\linewidth} \subcaption[]{\textcolor{orange}{The positive sample.}}
    \end{subfigure}
    \captionsetup{skip=3pt}
    \caption{The camera moves and the background is fixed in the sequence. Please answer if the moving direction of camera of A$\rightarrow$B
    is \textcolor{orange}{back} (\textcolor{olive}{back}), the moving direction of
    camera of B$\rightarrow$C is \textcolor{orange}{right} (\textcolor{olive}{left})
    and the moving direction of camera of C$\rightarrow$D is \textcolor{orange}{left}
    (\textcolor{olive}{right}). Please answer Yes, if all questions are correct.
    Otherwise, answer No.}
    \label{img:move}
\end{figure}

\textbf{Camera Movement (Ca. M.)} In this task, we keep the camera's angle fixed
and move it to a random position in space. We ensure that regardless of the
movement direction, the next position always retains visibility of a portion of the
previous space. Additionally, we restrict movement directions to front, back, left,
or right, as we have found that an excessive number of directions poses a significant
challenge for humans. This level of difficulty is not sufficiently reasonable
for current VLMs. \cref{img:move} shows the samples of this task.

\textbf{Object Heading Direction (Obj. H. D.)} The following three tasks focus
on the position of objects in 3D space. This task specifically addresses the detection
of the heading direction of a target object. We place a sheep in the image and
inquire about the heading direction of the target object. When querying the model,
we refer to the sheep block as the ``red block with a white head.''

\textbf{Object Movement Direction (Obj. M. D.)} This task centers on detecting the
movement of a target object, with particular attention to the object's
orientation. Specifically, we examine the movement of a sheep and query whether
the direction of its head aligns with the direction of its movement. We use an API
to set the orientation of the object, take a screenshot, and then call the API again
to move the object to the corresponding position. For negative samples, we either
shuffle the order of the images or set the orientation to be perpendicular to the
movement direction. So far, we have only tested movement along the x-axis.
During the dataset construction, we call an API to filter out any occluded images,
thereby ensuring the quality of the dataset. Samples for this task are shown in \cref{img:sheep_move_direction}.

\begin{figure}[t]
    \centering
    \begin{subfigure}
        [t]{0.225\textwidth}
        \centering
        \includegraphics[width=1\textwidth]{
            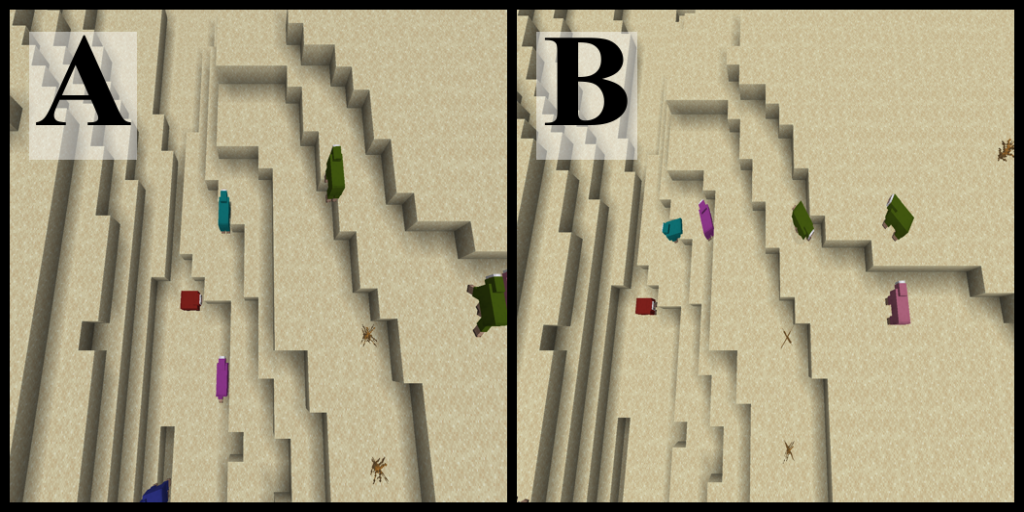
        } 
        \captionsetup{skip=3pt, width=1\linewidth, justification=centering}
        \subcaption[]{\textcolor{olive}{The negative sample.}}
    \end{subfigure}
    \begin{subfigure}
        [t] {0.225\textwidth}
        \centering
        \includegraphics[width=1\textwidth]{
            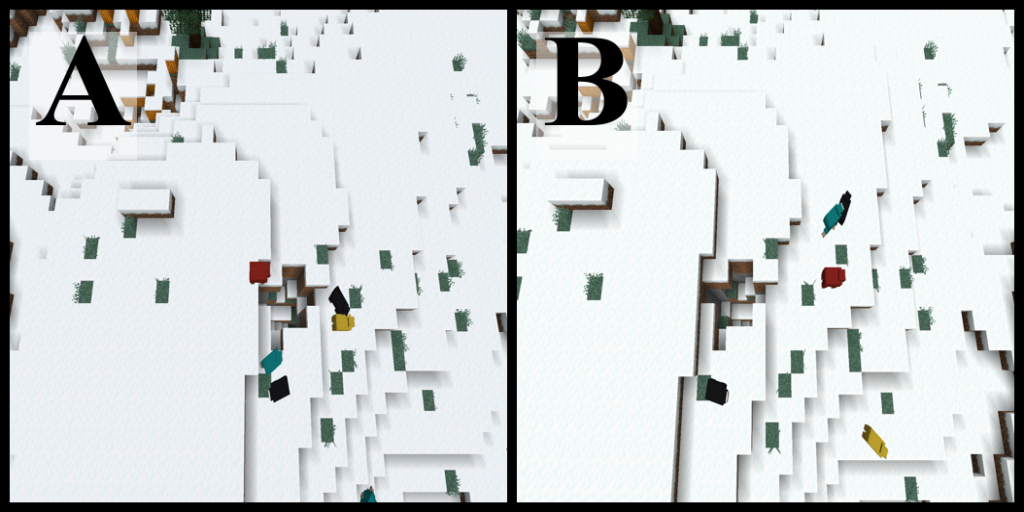
        }
        \captionsetup{skip=3pt, width=1\linewidth} \subcaption[]{\textcolor{orange}{The positive sample.}}
    \end{subfigure}
    \captionsetup{skip=3pt}
    \caption{The images have same background, but the camera moves. Each image has a orange sheep (orange
    block) which may move. Please answer if the orange sheep (orange block) has moved in the
    backgound between the two images. The answer is either Yes or No.}
    \label{img:sheep_move}
\end{figure}

\textbf{Object Movement (Obj. M.)} In this task, both the camera and the object
move. Each time the object moves, two screenshots are taken, with the camera also
moving each time. We then select two screenshots and ask the model whether the
object has moved. When selecting the images, we aim to ensure that the object's
absolute position in the images remains as similar as possible. This ensures that
the model must rely on spatial relationships to determine whether movement has occurred.
We query the VLMs to assess whether the object's position has changed between
the two images. To ensure a clear background in the screenshots, we maintain the
camera at a relatively high altitude above the ground while capturing the images.
As in the Object Movement Direction task, we ensure that the object remains
visible. Samples for this task are shown in \cref{img:sheep_move}.

For the tasks Object Heading Direction, Object Movement Direction, and Object Movement,
we have designed a \textbf{Clear} version. In the Clear version, we remove all other
objects from the image, leaving only the target object. Each task is constructed
with 200 samples, consisting of 100 positive samples and 100 negative samples. The
input image size for the models is $512 \times 512$ resolution for all tasks,
except for the Position Combination task, which uses a resolution of $1024 \times
1024$.

\begin{figure*}[t!]
    \centering
    \begin{subfigure}
        [t]{0.9\textwidth}
        \centering
        \includegraphics[width=\textwidth]{
            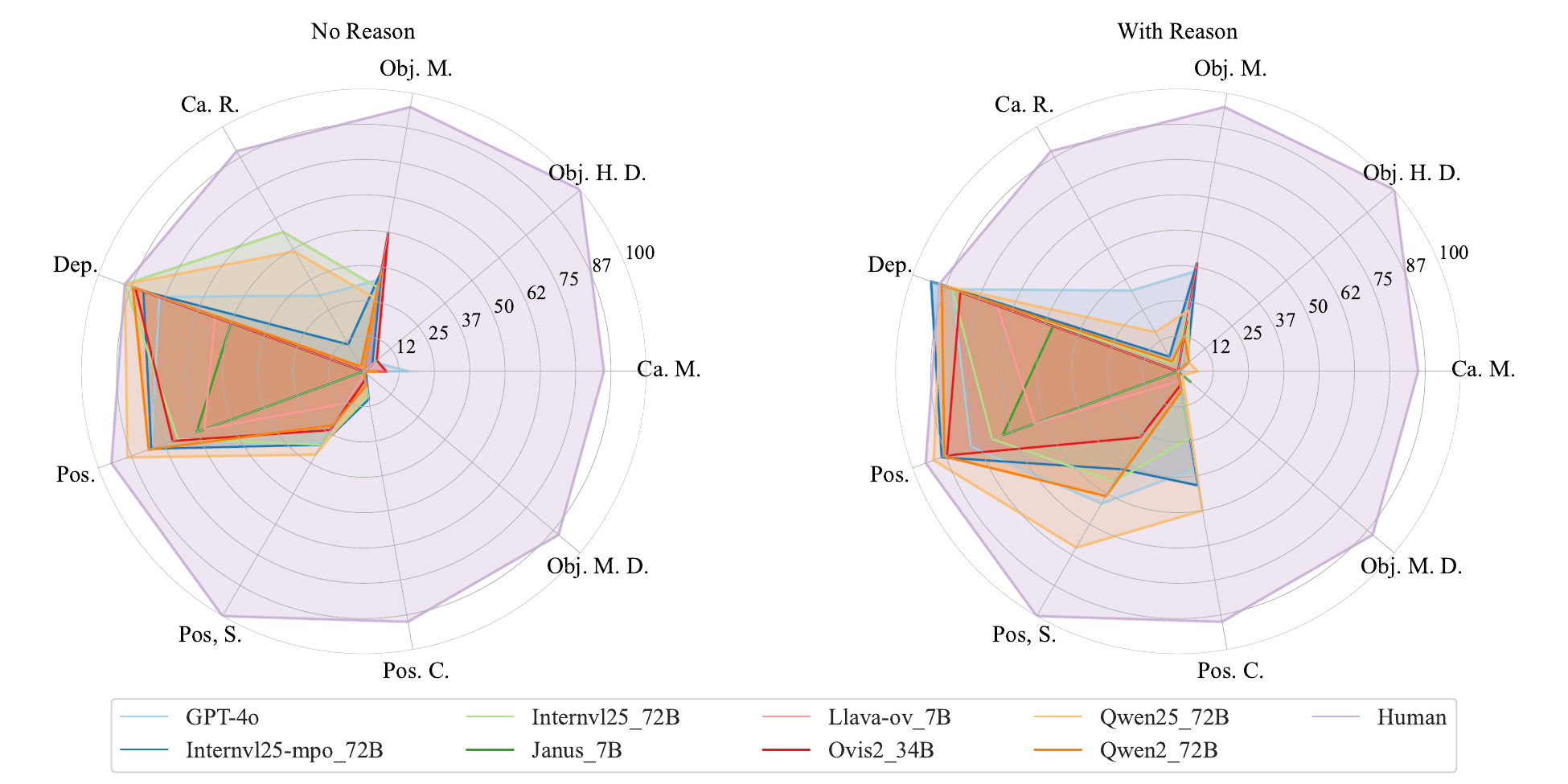
        } 
        \captionsetup{skip=0pt, width=1\linewidth, justification=centering}
    \end{subfigure}
    \captionsetup{skip=3pt}
    \caption{This chart shows the performance of the largest model in each
    series across different tasks.}
    \label{img:radar}
\end{figure*}

\subsection{Rationale for the Taxonomy}

We categorize spatial understanding into three primary dimensions: \textit{movement}, \textit{rotation}, and \textit{depth}. Within each category, we decompose the problem into granular sub-tasks to maximize coverage of spatial scenarios. Regarding the reviewer's concern about intermediate concepts like relative depth and containment, our existing benchmark addresses these through specific task designs:

\begin{itemize}
    \item \textbf{Relative Depth:} This requires a model to recognize objects and determine their sequential order along the depth axis. Our \textit{Position Sequence} task explicitly evaluates this capability. In this task, models must identify not just single images, but the sequential order of images. The cognitive capability required to process image sequences is analogous to the reasoning required for understanding the ``front-to-back'' ordering of objects in relative depth tasks.
    \item \textbf{Containment:} This involves understanding the inclusion relationship between one object and another. Our \textit{Object Movement} task addresses this by requiring the model to identify changes in the containment and relative positional relationship between the foreground object and the background environment. This effectively evaluates the core capabilities required for containment understanding.
\end{itemize}

Overall, our approach isolates specific spatial understanding capabilities through task decomposition. By modeling physical motion as the translation and rotation of rigid bodies (applied to both the object and the observer) and separating foreground from background, our combined tasks provide a comprehensive evaluation of spatial ability.

\subsection{Alignment with Real-World Perception}
We argue that our synthetic benchmark effectively proxies real-world spatial perception for two reasons:

\begin{enumerate}
    \item \textbf{Visual Realism in Static Tasks:} For tasks involving \textit{Depth} and \textit{Position}, we utilize Flux to generate samples. As demonstrated in \cref{img:depth}, these generated images possess a high degree of visual realism comparable to real-world photography.
    \item \textbf{Geometric Consistency in 3D Tasks:} For \textit{Movement} and \textit{Rotation} tasks, we design the samples based on the kinematic principles of the real world (i.e., translation and rotation of objects and observers). While the visual domain is synthetic (Minecraft), the spatial logic is identical to reality. To verify that the domain gap (visual recognition) is not the primary cause of failure, we included an ``IsMC'' control task in the last column of \cref{tab:direct}. The results show that models achieve near-perfect accuracy in identifying Minecraft objects. Since models can successfully recognize the objects and the spatial relationships follow real-world physics, we conclude that the performance on this benchmark is a valid indicator of real-world spatial reasoning capabilities.
\end{enumerate}

\subsection{Evaluation Metrics}
When calculating the score, for a given task $i$, the model's accuracy is
denoted as $p_{i}\in [0,100]$, and the corresponding task weight is $w_{i}$. The
score of task $i$ is defined as:

\begin{equation}
    s_{i}= 2( p_{i}- 50)\mathds{1}[p_{i}\ge 50],
\end{equation}

\noindent
$\mathds{1}[\cdot]$ is denoted as the indicator function. The overall score of a model
is computed as: $S =\sum_{i}s_{i}$.

%% file: sections/experiment.tex
\section{Experiment}

We evaluate multiple state-of-the-art VLMs, including the GPT-4o \cite{achiam2023gpt} series, Qwen-VL2
and 2.5 series \cite{bai2025qwen2}, InternVL2 series \cite{chen2024internvl}, Ovis series \cite{lu2024ovis}, Janus \cite{chen2025janus}, and Llava-OV series \cite{li2024llavaonevisioneasyvisualtask}. Due
to resource constraints, we utilize the AWQ-quantized versions, which are
officially provided, for models exceeding 40B parameters in our evaluation. 
We also evaluate some fine-tuned VLMs based on 3D datasets, including the llava-3D, SpaceQwen25, and the upgraded llava-3D version based on the Qwen2.5 model.
In
our evaluation, we do not employ complex reasoning techniques such as Chain of Thought
(CoT). However, we evaluated two distinct prompting strategies—a direct inquiry and a reasoning-based prompt. These results are presented in \cref{tab:direct} and \cref{tab:think}, respectively. We believe these prompting comparisons serve as a diagnostic intervention, reflecting how users typically interact with these models in practical applications.

For detecting positive and negative samples in the first prompting method, we
detect the word ``No.'' If this word is not present in the response, we classify the
output as a positive sample. This strategy is employed because we explicitly
prompt the model to output a ``Yes/No'' response. Additionally, we recruit
10 volunteers to participate in our tests. For each task, 40 samples are randomly
selected.

\begin{figure}[t]
    \centering
    \begin{subfigure}
        [t]{0.48\textwidth}
        \centering
        \includegraphics[width=\textwidth]{
            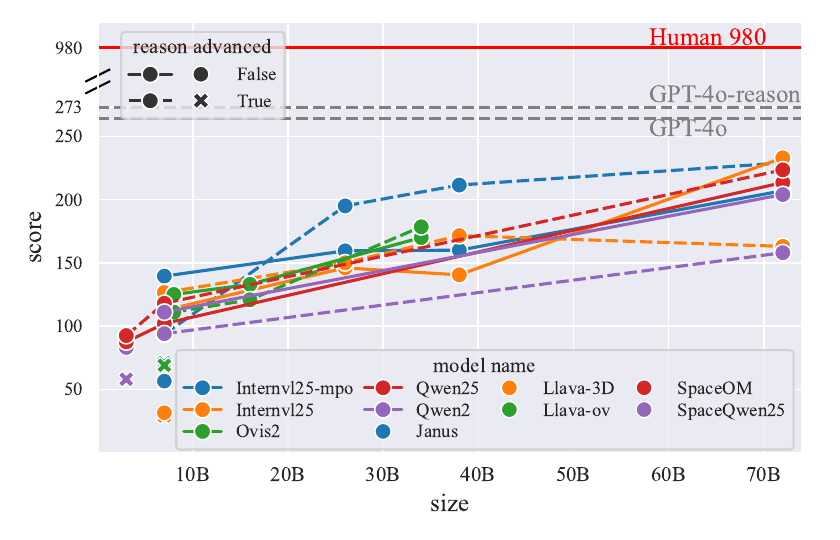
        } 
    \end{subfigure}
    \captionsetup{skip=3pt}
    \caption{The overall scores of different models on all tasks.}
    \label{img:summary}
\end{figure}

\subsection{The Gap of Total Performance between Models and Humans Is Significant}
\cref{img:summary} presents the overall scores of all models across all tasks.
From the results, it can be observed that as model size increases, the scores
improve for most models. All small models perform worse. Meanwhile, it can
be observed that reasoning advanced not necessarily improve its performance.
This suggests that reasoning might induce new hallucinations. The total score for
our task is 1050, while the highest score among the tested models is 272.5. However,
humans demonstrate near-perfect performance on all tasks, with a total score of 1050.
This indicates that there is still significant room for improvement in the models'
understanding of spatial relationships.

\cref{img:radar} presents a radar chart showing the performance of the largest
model in each series across different tasks. As we can see, it can be observed
that the models score close to zero on all 3D spatial understanding tasks
except for the task Depth. In terms of absolute position understanding, only the
simplest task Position achieves a score close to humans.




\begin{table*}
    \caption{\textsc{The Scores of Different Models on All Tasks (Answer after Reasoning).}}
    
    \centering
\begin{tabular}{llllllllll}
\toprule
 & Pos. & Pos. C. & Pos, S. & Dep. & Ca. M. & Ca. R. & Obj. H. D. (C) & Obj. M. (C) & Obj. M. D. (C) \\
\midrule
GPT-4-turbo & 46.0 & 9.0 & 36.0 & 68.0 & 3.0 & 5.0 & 0.0 (0.0) & 30.0 (18.0) & 7.0 (3.0) \\
GPT-4o-mini & 40.0 & 21.0 & 26.0 & 35.0 & \textbf{9.0} & \underline{19.0} & 4.0 (0.0) & 8.0 (12.0) & 11.0 (0.0) \\
GPT-4o & 78.0 & 35.0 & \underline{54.0} & \underline{85.0} & 2.0 & \textbf{33.0} & 1.0 (4.0) & \underline{36.0} (\textbf{46.0}) & 0.0 (\underline{7.0}) \\
\midrule Internvl25-mpo\_7B & 83.0 & 0.0 & 1.0 & 53.0 & 2.0 & 0.0 & \textbf{9.0} (0.0) & 4.0 (0.0) & 10.0 (0.0) \\
Internvl25-mpo\_26B & 81.0 & 26.0 & 34.0 & 65.0 & 0.0 & 9.0 & 1.0 (\textbf{12.0}) & 12.0 (27.0) & 1.0 (\textbf{17.0}) \\
Internvl25-mpo\_38B & 84.0 & 33.0 & 53.0 & 62.0 & 7.0 & 11.0 & 0.0 (\underline{11.0}) & 29.0 (19.0) & 0.0 (2.0) \\
Internvl25-mpo\_72B & 86.0 & \underline{41.0} & 40.0 & 78.0 & 5.0 & 5.0 & 7.0 (6.0) & \underline{36.0} (27.0) & 0.0 (0.0) \\
\midrule Internvl25\_7B & 79.0 & 0.0 & 3.0 & 59.0 & \underline{8.0} & 0.0 & 3.0 (2.0) & 0.0 (18.0) & \textbf{20.0} (5.0) \\
Internvl25\_26B & 70.0 & 11.0 & 15.0 & 57.0 & 0.0 & 10.0 & 0.0 (1.0) & 19.0 (22.0) & 16.0 (0.0) \\
Internvl25\_38B & 77.0 & 13.0 & 46.0 & 60.0 & 1.0 & 9.0 & 2.0 (4.0) & 17.0 (17.0) & \underline{17.0} (0.0) \\
Internvl25\_72B & 70.0 & 17.0 & 37.0 & 79.0 & 0.0 & 4.0 & 4.0 (7.0) & 27.0 (8.0) & 0.0 (3.0) \\
\midrule Janus\_7B & 65.0 & 0.0 & 2.0 & 49.0 & 0.0 & 0.0 & 0.0 (8.0) & 0.0 (0.0) & 5.0 (0.0) \\
\midrule Llava-3D\_7B & 16.0 & 1.0 & 0.0 & 12.0 & 0.0 & 0.0 & 3.0 (0.0) & 0.0 (4.0) & 2.0 (5.0) \\
Llava-ov\_7B & 48.0 & 9.0 & 5.0 & 67.0 & 0.0 & 0.0 & 0.0 (0.0) & 0.0 (0.0) & 0.0 (0.0) \\
\midrule Ovis2\_8B & 69.0 & 0.0 & 7.0 & 78.0 & 1.0 & 6.0 & \underline{8.0} (\underline{11.0}) & 0.0 (0.0) & 4.0 (4.0) \\
Ovis2\_16B & 66.9 & 0.8 & 30.0 & \textbf{87.0} & 0.0 & 17.0 & 0.0 (0.0) & 5.0 (0.0) & 6.0 (0.0) \\
Ovis2\_34B & 87.0 & 5.0 & 26.0 & 84.0 & 7.0 & 0.0 & 0.0 (2.0) & \textbf{60.0} (6.0) & 0.0 (0.0) \\
\midrule Qwen25\_3B & 74.0 & 0.0 & 2.0 & 71.0 & 0.0 & 12.0 & 0.0 (2.0) & 2.0 (0.0) & 0.0 (3.0) \\
Qwen25\_72B & \underline{89.0} & \textbf{47.0} & \textbf{72.0} & 80.0 & 1.0 & 11.0 & 5.0 (\underline{11.0}) & 21.0 (1.0) & 1.0 (5.0) \\
Qwen25\_7B & \textbf{93.0} & 3.0 & 18.0 & 55.0 & 3.0 & 0.0 & 0.0 (1.0) & 22.0 (6.0) & 0.0 (0.0) \\
\midrule Qwen2\_7B & 69.0 & 3.0 & 5.0 & 80.0 & 1.0 & 0.0 & 1.0 (5.0) & 5.0 (0.0) & 2.0 (0.0) \\
Qwen2\_72B & 83.0 & 5.0 & 51.0 & 60.0 & 2.0 & 4.0 & 0.0 (0.0) & 14.0 (\underline{36.0}) & 0.0 (0.0) \\
\midrule SpaceOM\_3B & 78.0 & 3.0 & 4.0 & 70.0 & 5.0 & 3.0 & 2.0 (0.0) & 2.0 (0.0) & 0.0 (0.0) \\
\midrule SpaceQwen25\_3B & 65.0 & 0.0 & 0.0 & 47.0 & 0.0 & 0.0 & 0.0 (1.0) & 0.0 (1.0) & 0.0 (0.0) \\
\midrule Human & 95.0 & 90.0 & 100.0 & 90.0 & 85.0 & 90.0 & 100.0 (100.0) & 95.0 (100.0) & 90.0 (95.0) \\
\bottomrule
\end{tabular}
\label{tab:think}
\end{table*}

\begin{table*}
    \caption{\textsc{The scores of different models on all tasks (Answer directly).}}
    \centering
\begin{tabular}{lllllllllll}
\toprule
 & Pos. & Pos. C. & Pos, S. & Dep. & Ca. M. & Ca. R. & Obj. H. D. (C) & Obj. M. (C) & Obj. M. D. (C) & IsMC\\
\midrule
GPT-4-turbo & 34.0 & 0.0 & 15.0 & 49.0 & 1.0 & 10.0 & 0.0 (0.0) & 5.0 (0.0) & 3.0 (0.0) & 100.0 \\
GPT-4o-mini & 36.0 & 1.0 & 1.0 & 41.0 & 0.0 & 9.0 & 0.0 (4.0) & 11.0 (12.0) & 0.0 (0.0) & \\
GPT-4o & 80.0 & \underline{10.0} & 29.0 & 77.0 & \textbf{16.0} & 31.0 & 5.0 (5.0) & 33.0 (\textbf{55.0}) & 0.0 (\textbf{16.0}) & \\
\midrule Internvl25-mpo\_7B & 80.0 & 5.0 & 10.0 & 85.0 & 0.0 & 0.0 & 4.0 (5.0) & 9.0 (20.0) & 1.0 (8.0)  & 100.0\\
Internvl25-mpo\_26B & 83.0 & 1.0 & 19.0 & 83.0 & 0.0 & 4.0 & 1.0 (1.0) & 29.0 (24.0) & 7.0 (0.0) \\
Internvl25-mpo\_38B & 81.0 & 4.0 & 16.0 & 85.0 & 6.0 & 10.0 & 1.0 (\underline{8.0}) & 19.0 (15.0) & 1.0 (5.0) \\
Internvl25-mpo\_72B & 80.0 & \underline{10.0} & \underline{31.0} & 83.0 & 3.0 & 11.0 & 4.0 (1.0) & \underline{36.0} (\underline{44.0}) & 1.0 (0.0) \\
\midrule Internvl25\_7B & 73.0 & 0.0 & 6.0 & 72.0 & 0.0 & 2.0 & 3.0 (2.0) & 18.0 (7.0) & 0.0 (5.0) & 96.0\\
Internvl25\_26B & 70.0 & 2.0 & 5.0 & 71.0 & 0.0 & 15.0 & 0.0 (0.0) & 27.0 (20.0) & \underline{9.0} (0.0) \\
Internvl25\_38B & 83.0 & 0.0 & 15.0 & 77.0 & 3.0 & 10.0 & 0.0 (3.0) & 14.0 (20.0) & 0.0 (3.0) \\
Internvl25\_72B & 70.0 & 9.0 & 30.0 & \textbf{90.0} & 0.0 & \textbf{57.0} & 0.0 (0.0) & 30.0 (42.0) & 0.0 (0.0) \\
\midrule Janus\_7B & 63.0 & 0.0 & 0.0 & 50.0 & 0.0 & 0.0 & 0.0 (0.0) & 0.0 (0.0) & 0.0 (0.0) & 100.0\\
\midrule Llava-3D\_7B & 17.0 & 2.0 & 0.0 & 6.0 & 4.0 & 0.0 & 5.0 (0.0) & 0.0 (4.0) & 2.0 (3.0) & 100.0\\
Llava-ov\_7B & 60.0 & 1.0 & 13.0 & 56.0 & 0.0 & 0.0 & 5.0 (\textbf{11.0}) & 0.0 (19.0) & 0.0 (0.0) & 100.0 \\
\midrule Ovis2\_8B & 68.0 & 7.0 & 2.0 & 76.0 & 0.0 & 12.0 & \textbf{8.0} (\underline{8.0}) & 7.0 (5.0) & 5.0 (0.0) & 100.0\\
Ovis2\_16B & 76.0 & 0.0 & 5.0 & 83.0 & 0.0 & 0.0 & 2.0 (1.0) & 26.0 (2.0) & \textbf{10.0} (\underline{10.0}) \\
Ovis2\_34B & 73.0 & 3.0 & 25.0 & \underline{86.0} & \underline{7.0} & 0.0 & \underline{6.0} (6.0) & \textbf{51.0} (1.0) & 0.0 (4.0) \\
\midrule Qwen25\_3B & 81.0 & 0.0 & 5.0 & 83.0 & 0.0 & 3.0 & 0.0 (0.0) & 0.0 (0.0) & 0.0 (0.0) & 100.0\\
Qwen25\_7B & \textbf{92.0} & 0.0 & 7.0 & 79.0 & 0.0 & 0.0 & 0.0 (1.0) & 6.0 (6.0) & 0.0 (0.0)\\
Qwen25\_72B & 89.0 & \textbf{17.0} & \textbf{34.0} & \textbf{90.0} & 5.0 & \underline{49.0} & 0.0 (2.0) & 26.0 (8.0) & 0.0 (0.0) \\
\midrule Qwen2\_7B & 72.0 & 0.0 & 1.0 & 85.0 & 0.0 & 7.0 & 3.0 (0.0) & 22.0 (0.0) & 0.0 (0.0) & 100.0\\
Qwen2\_72B & 81.0 & 5.0 & 22.0 & \underline{87.0} & 5.0 & 2.0 & 0.0 (\textbf{13.0}) & 36.0 (\underline{48.0}) & 0.0 (0.0) \\
\midrule SpaceOM\_3B & 82.0 & 3.0 & 1.0 & 81.0 & 0.0 & 7.0 & 0.0 (0.0) & 0.0 (0.0) & 0.0 (0.0) & 100.0\\
\midrule SpaceQwen25\_3B & \underline{91.0} & 0.0 & 9.0 & 66.0 & 0.0 & 0.0 & 0.0 (0.0) & 0.0 (0.0) & 0.0 (0.0) & 100.0\\
\midrule Human & 95.0 & 90.0 & 100.0 & 90.0 & 85.0 & 90.0 & 100.0 (100.0) & 95.0 (100.0) & 90.0 (95.0) & 100.0\\
\bottomrule
\end{tabular}
    \label{tab:direct}
\end{table*}

\subsection{Reasoning Does Not Always Improve Performance}

A simple strategy to enhance model performance is to prompt the model to provide
reasoning before delivering an answer. However, our results indicate that this
approach does not always yield improvements.

For simple tasks like identifying absolute position, the performance of
models with prior reasoning does not significantly differ from those that answer
directly. For instance, in the \textbf{Position} task, where models identify an
object's absolute location in an image, both Qwen2.5-72B and Qwen2.5-7B achieve
the highest score of 92, with minimal variation between the two prompting
strategies.

By contrast, for more complex tasks like the \textbf{Position Combination} task,
where models must determine the relative positioning of sub-images, reasoning
before answering significantly enhances performance. In this case, prior
reasoning improved performance by 31 points for Interval25-mpo-72B and by 43
points for Qwen2.5-72B. Other models also exhibit substantial performance
gains when prior reasoning is employed.

However, this trend does not hold across all challenging tasks. For certain
tasks, prior reasoning does not lead to improved performance. For example, in the
\textbf{Camera Rotation} task, which evaluates a model's ability to understand spatial
transformations due to camera rotation, models perform considerably worse when
required to generate reasoning first. The best-performing model, InternVL2.5-72B,
scores 57 without reasoning, but achieves a score of zero when prior reasoning is
included. A similar trend is observed in other tasks, such as \textbf{Object
Movement}. This suggests that, for some tasks, reasoning may introduce new hallucinations,
which can negatively affect accuracy.

\subsection{VLMs Are Incapable of 3D Spatial Understanding}

Our experiments demonstrate that the models struggle significantly with 3D spatial
understanding, particularly in tasks that require reconstructing a 3D scene or
tracking objects.

For example, in the \textbf{Camera Movement} task, where the model must reconstruct
a 3D scene from multiple images captured from different perspectives, nearly all
models scored zero, regardless of the prompting strategy used. The highest score,
achieved by GPT-4o, is only 16. Similarly, in the \textbf{Camera Rotation} task,
the results are comparable. The best-performing model, InternVL2.5-72B, achieves
a score of 57, while humans score over 90 points. These results suggest that VLMs
lack an intrinsic understanding of 3D transformations and struggle to integrate multi-view
information into a coherent spatial model.

In the \textbf{Object Movement} task, where models are required to track the motion
of an object relative to its background, performance remained poor. The best model,
GPT-4o, scores 55, and additional reasoning before answering does not result in
significant improvement. These results indicate that the models' approach to
object tracking is both inconsistent and unreliable.

Moreover, in the \textbf{Object Movement Direction} task, which involves determining
whether an object's heading direction aligns with its movement, we observe a
surprising phenomenon: larger models tend to perform worse. In some instances,
models even outperform their results from the simpler \textbf{Object Heading
Direction} task. However, this improvement does not necessarily reflect a
correct understanding of the images. For example, models frequently predict
directions such as “bottom-right,” despite the dataset containing only four
discrete directions (up, down, left, right). This suggests that VLMs are prone
to hallucinating movement patterns rather than accurately interpreting
structured motion cues.

\subsection{No Single Model or Method Is Universally Superior}

Our analysis reveals that no single model or reasoning strategy consistently outperforms
others across all spatial tasks. While some models excel in specific areas, they
can encounter difficulties in others, and the effectiveness of different
prompting methods varies depending on the task.

For instance, Qwen2.5-72B achieves the best performance in the \textbf{Position} and \textbf{Position
Combination} tasks, whereas InternVL2.5-72B leads in the \textbf{Camera Rotation}
and \textbf{Object Heading Direction} tasks. However, both models struggle
significantly with tasks involving object movement or complex 3D transformations.
Furthermore, the InternVL model, which is trained using MPO reinforcement
learning, surprisingly performs worse than its base version in some tasks. This
suggests that certain training techniques may not always generalize effectively
across all types of spatial reasoning problems.

We also evaluate several models including Llava-3D, SpaceOM, and SpaceQwen. SpaceQwen is derived from SpaceVLM~\cite{chen2024spatialvlm} and is fine-tuned using 3DSRBench datasets~\cite{ma20243dsrbench} based on Qwen-VL. SpaceOM incorporates additional datasets~\cite{wu2025spatialscore} beyond this foundation. Our experimental results demonstrate that these models show no significant improvement in performance. In fact, the performance of those models is degraded almost all tasks. This observation suggests that training on specific types of 3D datasets does not necessarily enhance a model's overall 3D spatial comprehension abilities.

Taken together, our findings suggest that VLMs still exhibit substantial gaps in
spatial understanding. While some models demonstrate strong performance in
isolated tasks, none possess a comprehensive and robust grasp of spatial reasoning
across a wide range of problems.

\subsection{Analysis of Model Failures}

Based on the results in \cref{tab:think} and \cref{tab:direct}, we attribute the near-zero scores and general failures to three primary causes:

\begin{enumerate}
    \item \textbf{Insufficient Granularity in Object Recognition:} This is particularly evident in smaller models. Analyzing the \textit{Pos.} and \textit{Depth} data—the simplest spatial tasks requiring only binary relational identification—some models (e.g., Llava-3D 7B) perform poorly. These models achieve poor performance in other tasks too. Besides, in the \textit{Movement} tasks, we provided a ``Clear'' version (denoted by (C)), where distractions were removed to leave only the essential objects. Most models showed improved performance on the Clear version, suggesting the lack of fine-grained object recognition.
    \item \textbf{Deficiencies in Logical Reasoning:} This is observed in the \textit{Pos. S} (Sequence) and \textit{Pos. C} (Combination) tasks. To succeed, models must build upon basic positioning capabilities by further understanding the sequential or compositional relationships between sub-images. The significant performance drop in these tasks compared to the basic \textit{Pos.} task indicates a failure in higher-order logical reasoning.
    \item \textbf{Incomprehensive Spatial Representations:} Models perform poorly on tasks involving intrinsic orientation, such as \textit{Obj. H. D.} (Heading Direction) and \textit{Obj. M. D.} (Movement Direction), as well as \textit{Camera Rotation}. This suggests that current VLMs lack a robust internal representation of 3D geometry, specifically regarding object orientation and the observer's viewpoint.
\end{enumerate}

%% file: sections/conclusion.tex
\section{Conclusion}

In this work, we present a fully synthetic dataset designed to evaluate spatial
understanding. Owing to its synthetic nature, the dataset mitigates the risk of
contamination from preexisting training data and can be easily extended to
incorporate more complex scenarios at low costs. Our dataset includes tasks targeting
both absolute spatial understanding and 3D spatial understanding. The 3D spatial
component is further divided into object-centric and camera-centric perspectives,
and spatial transformations are categorized into rotation and positional changes.
Based on this framework, we construct nine tasks and conduct evaluations on multiple
state-of-the-art VLMs. The results reveal that, in the domain of spatial understanding,
even the best-performing models only approach human-level performance on the two
simplest tasks, which remains the challenge for further VLMs. Additionally, we find that no single model consistently
outperforms others across all tasks. The simple strategy of prompting the model to
reason before answering actually leads to performance degradation in certain cases.
Furthermore, increasing model size, fine-tuning on specific 3D datasets and incorporating reinforcement learning do not
consistently enhance performance; in some cases, they even result in a decline.
These findings highlight persistent limitations in current VLMs' ability to perform
robust spatial reasoning, particularly in complex or 3D scenarios, and suggest several
promising directions for future research.